\documentclass[runningheads]{llncs}

\usepackage{eccv}

\usepackage{eccvabbrv}
\usepackage[table]{xcolor}
\usepackage{graphicx}
\usepackage{booktabs}
\usepackage{multirow}
\usepackage{wrapfig}

\usepackage{caption}
\usepackage[accsupp]{axessibility}  

\usepackage{hyperref}

\usepackage{orcidlink}

\usepackage{algorithm}
\usepackage{algpseudocode}

\begin{document}

\title{Action-Aligned Retrieval with Pairwise Multimodal Reranking for Text-Based Person Anomaly Search} 

\titlerunning{ActPair}

\author{
Thanh-Khoi Nguyen\inst{1,2}\textsuperscript{$\dagger$}\orcidlink{0009-0003-5879-451X}
\and
Thanh-Nhan Vo\inst{1,2}\textsuperscript{$\dagger$}\orcidlink{0009-0007-8403-1240}
\and
Trong-Thuan Nguyen\inst{1,2}\orcidlink{0000-0001-7729-2927}
\and
Minh-Triet Tran\inst{1,2}\orcidlink{0000-0003-3046-3041}
}

\authorrunning{Thanh-Khoi Nguyen et al.}

\institute{
University of Science, VNU-HCM, Vietnam
\and
Vietnam National University, Ho Chi Minh City, Vietnam\\
\email{\{ntkhoi,vtnhan,ntthuan\}@selab.hcmus.edu.vn}\\
\email{tmtriet@fit.hcmus.edu.vn}\\[2pt]
\textsuperscript{$\dagger$}These authors contributed equally.
}

\maketitle





\begin{abstract}
Text-based person anomaly search requires distinguishing individuals based on fine-grained, context-dependent behaviors rather than mere appearance. Existing methods struggle to capture these context-conditioned actions, frequently relying on isolated skeletal geometry, discarding raw query details during reformulation, or utilizing absolute pointwise scoring for multimodal verification. To address these limitations, we propose \textbf{ActPair}, a unified three-stage coarse-to-fine framework that combines action-aligned retrieval with pairwise multimodal reranking to bridge the pose-semantic gap. First, we fine-tune a vision-language model (VLM) with an action-aligned multi-task objective that encourages the representations to encode action-discriminative semantics. Second, we perform parallel late-fusion retrieval using the original query and a large language model (LLM)-generated context-grounded rewrite, retaining complementary details from both semantic views. Finally, we propose an efficient off-the-shelf reranking module that leverages a pivot-promote algorithm to perform direct pairwise visual comparisons, mitigating residual spatial and compositional ambiguities without the prohibitive inference costs of exhaustive evaluation. Extensive experiments demonstrate that our framework achieves the best results among the compared methods on the Pedestrian Anomaly Behavior (PAB) public test and and transfers effectively to an unseen, non-anomaly-specific dataset.
\end{abstract}

\section{Introduction}
Text-based person search retrieves individuals using natural language but is largely limited to routine activities, overlooking abnormal behaviors critical in surveillance. Recent work extends this to text-based person anomaly search, where individuals may share similar appearance and can only be distinguished by fine-grained actions described in the query. These actions are often entangled with implicit scene context, requiring representations that capture identity-independent behavior and context, rather than relying on appearance alone.

While recent advances have pushed the boundaries of text-based person retrieval, existing methods still exhibit critical limitations when applied to the highly nuanced task of person anomaly search. First, prior pose-aware frameworks (e.g., Cross-Modal Pose-aware (CMP) \cite{yang2025beyondwalking}) predominantly rely on isolated skeletal geometry as a supervisory signal for human action. However, in static image retrieval, where temporal dynamics are absent, skeletal posture is inherently ambiguous; a physical pose resolves into definitive action semantics only when explicitly coupled with its surrounding environmental context. Second, although recent methods explore query reformulation (e.g., MVR \cite{yuan2026mvr}, ReCQR \cite{lee2026recqr}, PlugIR \cite{lee2024interactive}, MRA \cite{yang2025minimizing}), they typically replace the original query entirely. They lack a parallel late-fusion mechanism capable of synergizing the unadulterated fine-grained details of the raw query with a canonical, context-grounded rewrite, thereby inadvertently discarding complementary semantic evidence. Third, global dense embeddings exhibit a well-documented vulnerability to the compositional binding problem. While highly proficient at recognizing isolated semantic concepts, they frequently fail to accurately bind these attributes to their correct subjects within a unified interaction graph. In anomaly search, distinguishing events rarely relies on the mere presence of objects, but rather on precise spatial interactions (e.g., "a man falling off a bicycle" versus "a man standing next to a fallen bicycle"). Consequently, cosine similarity metrics over compressed representations struggle with final fine-grained localization. To mitigate this, contemporary generative methods (e.g., AnomalyLMM \cite{ju2025anomalylmm}, IRRA \cite{jiang2023cross} reranking) introduce multimodal large language model (MLLM)-based reranking, but they typically rely on independent caption generation or absolute pointwise scoring. This isolated assessment forces models to maintain globally calibrated scales. However, it overlooks a fundamental intuition: assessing subtle visual differences is inherently easier and more reliable when directly comparing two highly similar candidates side-by-side, rather than attempting to compute an absolute relevance score for a single image in isolation. Finally, current architectures (e.g., SSDC \cite{xie2026bridging}, RaSa~\cite{yang2023rasa}, APTM \cite{yang2023towards}, CAMeL \cite{zhu2025camel}) remain fragmented, lacking a unified coarse-to-fine framework tailored for fine-grained person anomaly search.

To overcome these limitations, we propose \textbf{ActPair}, an action-aligned retrieval framework with pairwise multimodal reranking, designed as a unified three-stage coarse-to-fine pipeline to bridge the pose-semantic gap in text-based person anomaly search. First, we establish a robust dense retrieval foundation by fine-tuning a VLM via an action-aligned multi-task learning objective and cross-modal consistency to capture rich, context-conditioned action semantics. Second, to address the contextual ambiguity of free-form natural language, we employ a parallel late-fusion retrieval strategy that harmonizes the original query with an LLM-synthesized, context-grounded rewrite, maximizing candidate recall while preserving fine-grained raw details. Finally, to alleviate residual spatial and compositional ambiguities that global dense embeddings inherently struggle to capture, we utilize an efficient off-the-shelf multimodal reranking module. By using an MLLM as a direct pairwise comparator within the pivot-promote procedure, our framework performs targeted reranking over the top-ranked candidates rather than constructing a complete pairwise ordering. The procedure reduces the number of comparisons in the evaluated top-10 setting, although its worst-case complexity remains quadratic.

Our main contributions are as follows:
\begin{itemize}
    \item  We propose action-aligned multi-task representation learning for a VLM that combines contrastive image-text alignment with explicit action-tag supervision and cross-modal consistency.
    \item We propose context-grounded query rewriting with parallel late-fusion retrieval, enriching queries with action and scene cues while fusing complementary semantic views to improve recall.
    \item We propose an efficient off-the-shelf pairwise multimodal reranking with a pivot-promote strategy. Under a matched Qwen3.5-9B comparison, direct candidate comparison improves R@1 and mAP over pointwise scoring.
\end{itemize}

\label{sec:intro}
\section{Related Work}
\label{sec:rel}
\subsection{Pose-Aware and Action-Semantic Representation for Person Anomaly Search}

Conventional text-based person retrieval primarily learns cross-modal representations for identity-related appearance descriptions. Recent methods strengthen fine-grained alignment through implicit visual-textual relation reasoning~\cite{jiang2023cross}, relation-aware supervision, and sensitivity to meaningful textual variations~\cite{yang2023rasa}. However, they are mainly designed to distinguish identities by appearance rather than behaviors that determine normal or anomalous retrieval targets.

The PAB benchmark introduces person anomaly search, where visually similar pedestrians must be distinguished by their behaviors~\cite{yang2025beyondwalking}. Its CMP framework incorporates human-pose patterns and identity-based hard negatives to capture structural behavioral variations. Nevertheless, similar skeletal configurations may correspond to different actions. SSDC formalizes this limitation as the \emph{pose-semantic gap} and combines pose-aware retrieval with MLLM-based semantic verification~\cite{xie2026bridging}, although semantic reasoning is applied mainly after the initial structure-guided stage.

Studies in human-object interaction and action recognition further show that action semantics depend on interacting objects and scene context, rather than body geometry alone~\cite{wang2019deep,luo2024discovering}. At the same time, background context must be modeled carefully to avoid spurious scene shortcuts~\cite{li2023mitigating}. Motivated by these observations, our approach directly optimizes the SigLIP2 embedding space~\cite{tschannen2025siglip} using explicit action supervision and cross-modal action consistency, enabling context-conditioned action representations before retrieval.

\subsection{Multimodal Reranking for Fine-Grained Retrieval}

Learning-to-rank methods construct rankings via pointwise, pairwise, or listwise relevance modeling~\cite{burges2005learning,cao2007learning}. Recent language and multimodal models extend these approaches beyond dense-encoder similarity scores.

In text-based person anomaly search, SSDC verifies each retrieved image with an MLLM and fuses image-specific semantic evidence with pose-aware retrieval~\cite{xie2026bridging}, while AnomalyLMM independently interprets candidates through masked action and color descriptions before LLM-based ranking~\cite{ju2025anomalylmm}. Both process candidates independently before assessing relative relevance. Pairwise ranking prompting instead directly compares candidates, avoiding independently calibrated relevance scores~\cite{qin2024large}, but exhaustive pairwise comparison incurs quadratic multimodal inference cost~\cite{zhuang2024setwise}. We therefore compare only a small top-ranked candidate pool using a pivot-promote procedure, enabling fine-grained relative reasoning without exhaustive all-pairs evaluation.

\subsection{Query Reformulation and Multi-View Retrieval}

Recent vision-language retrieval methods reformulate or enrich textual queries at inference time to reduce ambiguity. PlugIR converts dialogue-form contexts into self-contained queries and generates candidate-aware clarification questions for interactive text-to-image retrieval~\cite{lee2024interactive}. ReCQR similarly rewrites long or underspecified conversational histories into concise, semantically complete queries~\cite{lee2026recqr}. Relevance-feedback methods instead refine the query using top-ranked images, synthetic captions, or learned multimodal feedback~\cite{khaertdinov2026little}.

MVR mitigates expression drift by generating keyword-preserving, semantically equivalent reformulations and aggregating their textual features. It also reformulates VLM-generated gallery descriptions for visual-semantic compensation~\cite{yuan2026mvr}. In contrast, our method generates a single action-scene-grounded rewrite for person anomaly search while retaining the original caption as a complementary retrieval view. The rewrite emphasizes behavioral context, whereas the original preserves instance-specific details. We combine these two views through late-fusion retrieval rather than aggregating multiple reformulations.

\begin{figure}[!t]
    \centering
    \includegraphics[width=\linewidth]{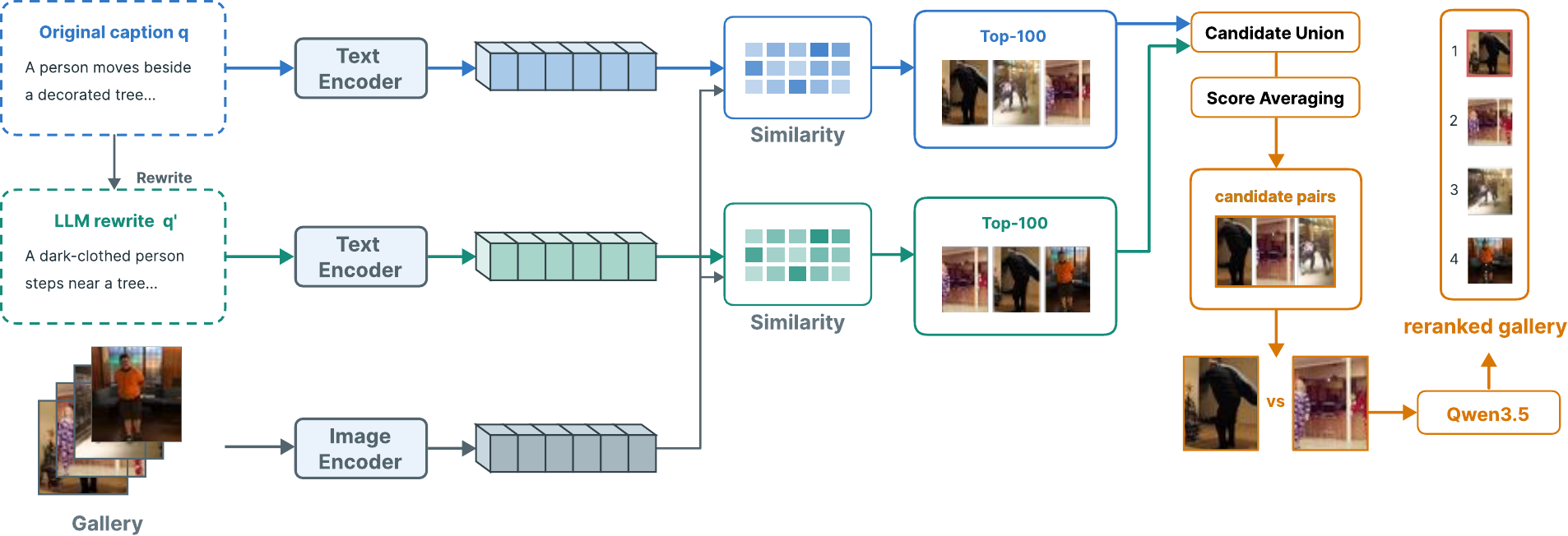}
    \caption{Overview of \textbf{ActPair}. An action-aligned dual encoder retrieves candidates from the original query and a context-grounded rewrite. Their rankings are fused before pairwise multimodal reranking with pivot-promote.}
    \label{fig:framework}
\end{figure}

\vspace{-2mm}
\section{Proposed Method}
\label{sec:method}

\vspace{-2mm}
\subsection{Problem Formulation}

Let $\mathcal{G} = \{I_i\}_{i=1}^{N}$ denote an image gallery of $N$ candidate pedestrian images, and let $Q$ be a natural language query describing a target pedestrian's appearance and behavior. The goal of text-based person anomaly search is to retrieve the image as formulated in Eq.~\eqref{eq:problem1}.
\begin{equation}
I^* = \arg\max_{I \in \mathcal{G}} \; \mathrm{sim}\big(Q, I\big),
\label{eq:problem1}
\end{equation}

such that $I^*$ depicts the pedestrian engaged in the exact normal or anomalous behavior specified by $Q$. We seek a ranking function as defined in Eq.~\eqref{eq:problem}.
\begin{equation}
    \mathcal{F}(Q, \mathcal{G}) \rightarrow \hat{\mathcal{G}},
    \label{eq:problem}
\end{equation}
that produces a sorted candidate list $\hat{\mathcal{G}}$ minimizing the rank position of the ground-truth image $I^*$. Given the scale of real-world galleries, $\mathcal{F}$ is instantiated in practice as a
composition of a high-recall retrieval stage over $\mathcal{G}$ followed by a fine-grained scoring stage restricted to a small candidate subset, though the formulation above remains agnostic to this decomposition.

\textbf{Overview.} As illustrated in Fig.~\ref{fig:framework}, \textbf{ActPair} consists of
three stages: action-aligned representation learning, dual-view retrieval
using the original and context-grounded queries, and pairwise multimodal
reranking with pivot-promote.

\vspace{-2mm}
\subsection{Offline Action and Scene Tag Extraction}
\label{sec:tag-extraction}

\vspace{-2mm}
We utilize the action and scene tags provided by training set PAB~\cite{yang2025beyondwalking} as the basis for constructing a fixed behavioral vocabulary. Let $\mathcal{R}_{act}$ and $\mathcal{R}_{scene}$ denote the sets of raw, free-form action and scene tags in the training annotations, and let $c(r)$ denote the occurrence count of a raw tag $r$. Owing to the heavy-tailed distribution of tag frequency, we retain only the top-$K$ most frequent tags in each category according to Eq.~\eqref{eq:topk-tags}.
\begin{equation}
    \mathcal{R}_{act}^{K} = \operatorname{top\text{-}}K_{r \in \mathcal{R}_{act}} \; c(r),
    \qquad
    \mathcal{R}_{scene}^{K} = \operatorname{top\text{-}}K_{r \in \mathcal{R}_{scene}} \; c(r).
    \label{eq:topk-tags}
\end{equation}

This together accounts for the large majority of dataset occurrences while discarding rare, noisy, near-duplicate phrasings. The retained tags, together with their frequencies, are provided to a LLM $\mathcal{M}_{tag}$, which induces a fixed set of $n_{act}$ and $n_{scene}$ canonical categories and produces a mapping in Eq.~\eqref{eq:tag-mapping}.
\begin{equation}
    \phi: \mathcal{R}_{act}^{K} \cup \mathcal{R}_{scene}^{K}
\;\rightarrow\;
\mathcal{T}_{act} \cup \mathcal{T}_{scene},
\label{eq:tag-mapping}
\end{equation}

Mitigating synonymy and surface-level variation between raw tags that describe the same underlying action or scene. The resulting lookup tables $\mathcal{T}_{act}$ and $\mathcal{T}_{scene}$ standardize the raw
annotations into a fixed vocabulary, which supplies the ground-truth labels $y_{act}$ for action supervision (Sec.~\ref{sec:representation-learning}) and the tag vocabulary used for query rewriting
(Sec.~\ref{sec:rewrite-fusion}).

\begin{figure}[!t]
    \centering
    \includegraphics[width=\linewidth]{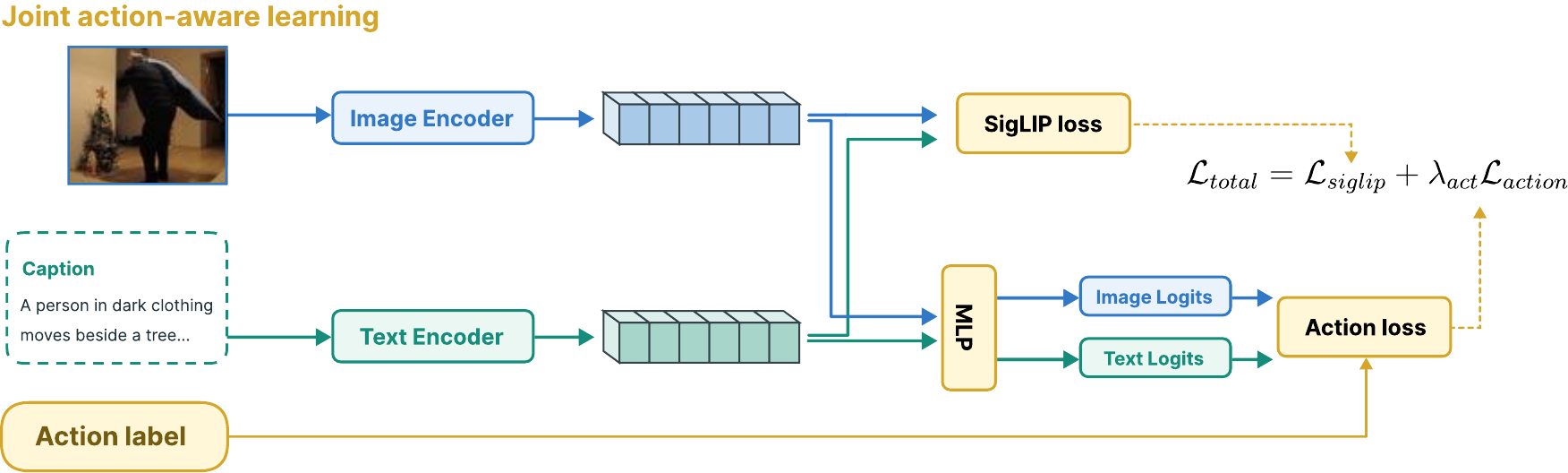}
    \caption{Action-aligned representation learning. Image and text embeddings are optimized by the SigLIP2 objective, shared action supervision, and cross-modal action consistency.}
    \label{fig:multi-task}
\end{figure}
\subsection{Multi-Task Representation Learning}
\label{sec:representation-learning}








Figure~\ref{fig:multi-task} illustrates the joint image-text alignment and action-supervision architecture. The effectiveness of our overall framework depends on the quality of the underlying dense retrieval model, which must produce a highly accurate initial candidate pool before any downstream contextualization or reranking can take place. Relying on an out-of-the-box VLM often yields suboptimal results, as such models can exhibit insufficient sensitivity when evaluating images with high pixel overlap (e.g., identical backgrounds and subjects), failing to distinguish subtle kinematic variations. To optimize the base SigLIP2 encoder for fine-grained anomaly retrieval, we propose a multi-task fine-tuning objective. Let $E_{img}$ and $E_{txt}$ represent the normalized image and text embeddings, and $s(i, j) = E_{img}^{(i)} \cdot E_{txt}^{(j)}$ denote their cosine similarity. The total training objective is formulated as Eq.~\eqref{eq:loss}.
\begin{equation}
    \label{eq:loss}
    \mathcal{L}_{total} = \mathcal{L}_{siglip} + \lambda_{action} \mathcal{L}_{action}
\end{equation}  
where $\mathcal{L}_{siglip}$ is the native SigLIP2 pairwise sigmoid loss utilized to maintain global, memory-efficient image-text alignment across the batch.

\noindent\textbf{Action Supervision and Semantic Consistency:} To reduce the visual encoder's reliance on dominant background cues, we propose an auxiliary classification head over predefined action tags. This encourages the model to allocate representational capacity to action-discriminative visual evidence beyond scene context alone. Let $p_{\mathrm{img}}=\operatorname{softmax}(W_{\mathrm{act}}E_{\mathrm{img}})$ and $p_{\mathrm{txt}}=\operatorname{softmax}(W_{\mathrm{act}}E_{\mathrm{txt}})$ denote the predicted distributions over the action-tag set $\mathcal{T}_{\mathrm{act}}$, produced from the image and text embeddings by the shared classification head $W_{\mathrm{act}}$. We define the action supervision objective as Eq.~\eqref{eq:action-loss}.
\vspace{-2mm}
\begin{subequations}
\begin{align}
    \mathcal{L}_{\mathrm{cons}}
    &=
    \frac{1}{2}
    \left[
    D_{\mathrm{KL}}\!\left(
    p_{\mathrm{img}} \,\Vert\, p_{\mathrm{txt}}
    \right)
    +
    D_{\mathrm{KL}}\!\left(
    p_{\mathrm{txt}} \,\Vert\, p_{\mathrm{img}}
    \right)
    \right],
    \\
    \mathcal{L}_{\mathrm{action}}
    &=
    \frac{1}{2}
    \left[
    \operatorname{CE}(p_{\mathrm{img}},y_{\mathrm{act}})
    +
    \operatorname{CE}(p_{\mathrm{txt}},y_{\mathrm{act}})
    \right]
    +
    \mathcal{L}_{\mathrm{cons}} .
    \label{eq:action-loss}
\end{align}

\end{subequations}
Here, $y_{\mathrm{act}}$ is the class index, equivalently a one-hot target, of the canonical action assigned to the target pedestrian. The auxiliary task is therefore formulated as single-label multi-class classification over $\mathcal{T}_{\mathrm{act}}$. The consistency term $\mathcal{L}_{\mathrm{cons}}$ is the symmetric Kullback-Leibler (KL) divergence between $p_{\mathrm{img}}$ and $p_{\mathrm{txt}}$. It encourages both modalities to agree on the action semantics of each image-text pair, thereby reducing their reliance on spurious background correlations.

\subsection{Query Rewriting and Parallel Late-Fusion Retrieval}
\label{sec:rewrite-fusion}

In static text-to-image retrieval, the original query $Q_{orig}$ often lacks structural alignment, leaving actions ambiguous without temporal dynamics. To reduce this cross-modal ambiguity prior to retrieval, we propose an offline context-grounded query enrichment mechanism that explicitly conditions actions on the surrounding environment.

\noindent\textbf{Query Rewriting.} Given a query $Q_{orig}$, we assign it the action and scene tags $\tau_{act}, \tau_{scene}$ - extracted via LLM inference over the canonical vocabulary - most relevant to its content from the canonical vocabulary obtained during offline tag extraction (Sec.~\ref{sec:tag-extraction}), and use an LLM $\mathcal{M}_{llm}$ to rewrite $Q_{orig}$ into a fixed template:
\begin{quote}
    \textit{``The image shows [person description] wearing [clothing]
    [action]. The background features [scene details].''}
\end{quote}

This process produces the rewritten query  $Q_{rew} = \mathcal{M}_{llm}(Q_{orig} \oplus \tau_{act} \oplus \tau_{scene})$ where $\tau_{act} \in \mathcal{T}_{act},$ and $ \tau_{scene} \in \mathcal{T}_{scene}$. By explicitly injecting action and scene tags, $Q_{rew}$ produces a context-grounded textual representation to ground the described behavior within its specific scene, mitigating the inherent ambiguities of the
original caption.

\noindent\textbf{Parallel Late-Fusion Retrieval.} While the enriched query $Q_{rew}$ mitigates spatial ambiguity, we treat the original caption $Q_{orig}$ and its rewrite $Q_{rew}$ as two complementary semantic views of the same retrieval intent. Specifically, $Q_{rew}$ provides the necessary structural grounding of action and background, whereas $Q_{orig}$ retains the raw, unadulterated fine-grained constraints (e.g., specific clothing colors or secondary object interactions) that might be
inadvertently altered or omitted during LLM synthesis. To leverage both perspectives, we employ a parallel late-fusion retrieval strategy.

Let $E$ denote the fine-tuned dual-encoder (SigLIP2~\cite{tschannen2025siglip}) that projects both textual queries and images into a shared $d$-dimensional embedding space. We independently execute two retrieval branches across the entire image gallery $\mathcal{G}$. The first branch computes cosine similarity scores $S_{orig}$ using $Q_{orig}$, while the second computes $S_{rew}$ using $Q_{rew}$. From these independent distributions, we extract the top-$K$ candidate subsets, denoted $\mathcal{C}_{orig}$ and $\mathcal{C}_{rew}$. To maximize candidate recall before the computationally expensive reranking stage, we construct a union candidate pool $\mathcal{C}_{union} = \mathcal{C}_{orig} \cup \mathcal{C}_{rew}$, preserving the visual evidence retrieved by either semantic view. For each candidate image $I_i \in \mathcal{C}_{union}$, we compute a fused similarity score $S_{final}(I_i)$ as the arithmetic mean of its corresponding scores from both branches, as formulated in Eq.~\eqref{eq:score}.
\begin{equation}
    S_{final}(I_i) = \frac{1}{2}\Big(S_{orig}(I_i) + S_{rew}(I_i)\Big).
    \label{eq:score}
\end{equation}

Finally, the union pool $\mathcal{C}_{union}$ is re-sorted in descending order based on $S_{final}$, and a highly probable top-$M$ candidate subset $\mathcal{C}_M$ (where $M < K$) is isolated and forwarded to the multimodal reranking module in Sec.~\ref{sec:reranking}.

\subsection{Pairwise Multimodal Reranking}
\label{sec:reranking}

\begin{algorithm}[t]
\caption{Pivot-Promote Pairwise Reranking}
\label{alg:pivot-promote}
\begin{algorithmic}[1]
\Require Query $Q$; ranked candidates
$\mathcal{R}=[I_1,\ldots,I_M]$; pairwise comparator
$\mathcal{C}(Q,I_a,I_b)\in\{a,b,\texttt{tie}\}$
\Ensure Reranked candidates
\Function{PivotPromote}{$Q,\mathcal{R}$}
    \If{$|\mathcal{R}|\leq 1$}
        \State \Return $\mathcal{R}$
    \EndIf

    \State $I_a \gets \mathcal{R}[1]$, $\mathcal{I}^{+} \gets [\,]$,
           $\mathcal{I}^{-} \gets [\,]$

    \For{$I_b \in \mathcal{R}[2{:}]$}
        \State \textbf{append} $I_b$ to $\mathcal{I}^{+}$ if
        $\mathcal{C}(Q,I_a,I_b)=b$; otherwise to $\mathcal{I}^{-}$
    \EndFor

    \State \Return
    $\Call{PivotPromote}{Q,\mathcal{I}^{+}}
    \oplus [I_a] \oplus \mathcal{I}^{-}$
\EndFunction
\end{algorithmic}
\end{algorithm}


\noindent\textbf{Pairwise Comparison as a Verification Signal.} Rather than prompting a VLM to assign an absolute scalar score to each image-text pair, we cast reranking as a sequence of pairwise comparisons. We leverage Qwen3.5~\cite{qwen35blog} as a pairwise comparator $\mathcal{M}_{cmp}$, and empirically validate this design choice against a pointwise scoring baseline in Sec.~\ref{ab:pair_wise}.

Given the original query caption $Q_{orig}$ and a candidate pair $(I_a, I_b)$, $\mathcal{M}_{cmp}$ is prompted to compare the two images along a fixed set of fine-grained criteria, namely the primary action,
target person, clothing, object interaction, secondary people, scene and background, and spatial relationships, together with any explicit contradictions or missing details relative to $Q_{orig}$. The comparator returns a single decision as defined in Eq.~\eqref{eq:compare}.
\begin{equation}
w = \mathcal{M}_{cmp}(Q_{orig}, I_a, I_b) \in \{a, b, \mathrm{tie}\},
\label{eq:compare}
\end{equation}

indicating whether $I_a$, $I_b$, or neither is better supported by the query.

\noindent\textbf{Pivot-Promote Reranking.} The complete reranking procedure is summarized in Algorithm~\ref{alg:pivot-promote}. Operating on the highly probable candidate subset $\mathcal{C}_{M}$ (where $M < K \ll N$) retrieved from the late-fusion stage, we employ a pivot-promote reranking procedure that recursively processes candidates preferred to the current pivot while preserving the prior order of the remaining candidates. An initial pivot candidate $I_a \in \mathcal{C}_M$, determined by the prior dense retrieval score, is compared against each remaining candidate $I_j \in \mathcal{C}_M \setminus \{I_a\}$ using $\mathcal{M}_{cmp}$, partitioning $\mathcal{C}_M \setminus \{I_a\}$ into the two subsets defined in Eq.~\eqref{eq:partition}.
\begin{equation}
    \mathcal{C}_{better} = \{I_j : w = b\}, \qquad
    \mathcal{C}_{worse} = \{I_j : w \in \{a, \mathrm{tie}\}\}.
    \label{eq:partition}
\end{equation}

If $|\mathcal{C}_{better}| > 1$, the algorithm recurses on $\mathcal{C}_{better}$, selecting a new pivot from within this subset and repeating the partitioning step; candidates in $\mathcal{C}_{worse}$ retain their relative order from the prior stage. The final ranking is obtained by concatenating the recursively sorted $\mathcal{C}_{better}$, the pivot $I_a$, and $\mathcal{C}_{worse}$, which rapidly elevates the ground-truth image toward the top of the ranking. Since pivot-promote recursively processes only candidates preferred to the current pivot, it has expected linear comparison complexity under a uniformly distributed pivot-rank assumption, while its worst-case complexity remains quadratic. In practice, the candidates are already approximately ordered by the preceding retrieval stages, typically yielding smaller recursive subsets and fewer comparisons.

\section{Experiment}
\subsection{Experimental Setup}

\textbf{Datasets and Baselines.} To comprehensively evaluate our framework, we utilize the PAB dataset~\cite{yang2025beyondwalking} and the RSTPReid benchmark~\cite{zhu2021dssl}. PAB is specifically constructed for person anomaly search, containing over 1 million image-text pairs spanning 1,600 anomalous and 1,000 normal behaviors. We use its augmented version from the AI City Challenge Track 4\footnote{\url{https://www.aicitychallenge.org/2026-track4/}} to test model robustness in real-world surveillance scenarios. Additionally, we use RSTPReid (20,505 images of 4,101 persons across 15 cameras) to validate generalization in non-anomaly-specific environments featuring complex transformations. We compare our approach against the PAB-specific CMP baseline~\cite{yang2025beyondwalking}, which explicitly integrates pose patterns to distinguish actions. Furthermore, we benchmark against general VLM (CLIP~\cite{radford2021clip}, X-VLM~\cite{zeng2021xvlm}) and state-of-the-art appearance-matching architectures (IRRA~\cite{jiang2023cross}, APTM~\cite{yang2023towards}, RaSa~\cite{yang2023rasa}, WoRA~\cite{sun2025filteringwora}, MRA~\cite{yang2025minimizing}, and CAMeL~\cite{zhu2025camel}). These models are primarily developed for appearance-oriented person retrieval and may therefore be less sensitive to fine-grained behavioral distinctions.

\textbf{Implementation Details.} We use Qwen3.5-9B~\cite{qwen35blog} to canonicalize the raw annotations into action and scene tags. The retriever is initialized from SigLIP2-Patch16-256~\cite{tschannen2025siglip} and fine-tuned on the PAB training set using the proposed objective with \(\lambda_{\mathrm{action}}=0.2\). At inference, we retrieve the top-\(K=100\) candidates for each query view, fuse their scores, and retain the top-\(M=10\) candidates for pairwise reranking with Qwen3.5-9B. All Qwen inference uses deterministic decoding with temperature \(=0\) and seed \(=0\).

\begin{table*}[!t]
\centering
\caption{Quantitative comparison on the original PAB test protocol under training-free and PAB fine-tuning settings.}
\label{tab:PAB_main_exp}

\begin{tabular}{llcccc|cccc}
\toprule
&& \multicolumn{4}{c}{\textbf{Training-free}} 
& \multicolumn{4}{c}{\textbf{Fine-tuned on PAB}} \\
\cmidrule(lr){3-6}
\cmidrule(lr){7-10}

\textbf{Method}
& \textbf{Venue}
& \textbf{R@1}
& \textbf{R@5}
& \textbf{R@10}
& \textbf{mAP}
& \textbf{R@1}
& \textbf{R@5}
& \textbf{R@10}
& \textbf{mAP} \\
\midrule

MRA~\cite{yang2025minimizing} 
& PR'26 
& 9.91 & 23.66 & 31.45 & 17.15
& 70.53 & 94.69 & 97.47 & 81.59 \\

RaSa~\cite{yang2023rasa} 
& IJCAI’23
& 21.74 & 27.30 & 27.96 & 24.35
& 80.79 & 98.89 & 99.65 & 89.20 \\

WoRA~\cite{sun2025filteringwora}
& WWW'25
& 22.25 & 45.91 & 53.54 & 33.39
& 74.47 & 96.82 & 98.48 & 84.60 \\

APTM~\cite{yang2023towards}
& MM'23
& 22.90 & 45.80 & 52.38 & 33.56
& 72.14 & 95.30 & 97.17 & 82.78 \\

CAMeL~\cite{zhu2025camel}
& TIFS'25
& 24.47 & 50.00 & 58.75 & 36.75
& 74.30 & 96.79 & 98.84 & 84.20 \\

IRRA~\cite{jiang2023cross}
& CVPR'23
& 30.59 & 59.61 & 68.91 & 44.41
& 76.39 & 97.62 & 99.14 & 86.33 \\

CLIP~\cite{radford2021clip}
& ICML'21
& \underline{47.57}
& \underline{81.55}
& \underline{89.03}
& \underline{62.73}
& 77.60
& 98.84
& \underline{99.75}
& 87.35 \\

X-VLM~\cite{zeng2021xvlm}
& ICML'22
& \textbf{71.94}
& \textbf{97.78}
& \textbf{98.99}
& \textbf{83.96}
& 81.95
& 98.84
& 99.19
& 89.86 \\

\midrule


CMP~\cite{yang2025beyondwalking}
& ICCV'25
& -- & -- & -- & --
& 84.93
& \underline{99.09}
& \underline{99.75}
& 91.66 \\

SSDC~\cite{xie2026bridging}
& ACL'26
& -- & -- & -- & --
& \underline{87.21}
& \underline{99.09}
& \underline{99.75}
& \underline{92.87} \\

\rowcolor{blue!5}
\textbf{ActPair (Ours)}
& ECCVW'26
& -- & -- & -- & --
& \textbf{88.62}
& \textbf{99.75}
& \textbf{99.85}
& \textbf{93.97} \\

\bottomrule
\end{tabular}
\end{table*}

\textbf{Evaluation Metrics.} We adopt standard evaluation metrics widely used in cross-modal retrieval tasks to assess the model's performance. Specifically, we report Recall@K (R@1, R@5, R@10) and mean Average Precision (mAP). Recall@K measures the percentage of natural language queries where the ground-truth matching image is successfully located within the top-K retrieved candidates. The mAP metric evaluates the overall area under the precision-recall curve, providing a comprehensive assessment of the model's global ranking quality across all queries.

\subsection{Quantitative Results}

\textbf{Main Results on PAB.}
ActPair achieves the best results among the compared published baselines on the original PAB test under the reported evaluation protocol, obtaining 88.62\% R@1 and 93.97\% mAP, as shown in Table~\ref{tab:PAB_main_exp}. In the training-free setting without domain-specific adaptation, conventional appearance-oriented architectures, including IRRA~\cite{jiang2023cross}, RaSa~\cite{yang2023rasa}, and APTM~\cite{yang2023towards}, struggle to capture fine-grained behavioral anomalies, with the strongest of these methods reaching only 44.41\% mAP. Foundational VLMs perform considerably better, with X-VLM~\cite{zeng2021xvlm} achieving 83.96\% mAP, yet still remain below methods explicitly adapted to the PAB domain.

Under the PAB fine-tuning setting, ActPair consistently outperforms both general-purpose person retrieval methods and PAB-specific baselines. Compared with the pose-aware CMP~\cite{yang2025beyondwalking}, our method improves R@1 and mAP by 3.69 and 2.31 percentage points, respectively. It also surpasses the stronger cascade-based SSDC~\cite{xie2026bridging} baseline by 1.41 points in R@1 and 1.10 points in mAP. Notably, the improvements are most pronounced at rank 1 and in overall ranking quality, while R@5 and R@10 are already close to saturation. This suggests that ActPair primarily improves the fine-grained ordering of highly similar candidates, rather than merely increasing coarse candidate recall.

\begin{table}[t]
\centering
\caption{
Comparison on RSTPReid under supervised fine-tuning and
cross-dataset transfer. In the zero-shot setting, IVT, IRRA, CFine,
APTM, and RaSa are trained on ICFG-PEDES, while CMP and Ours are
trained on PAB. ``--'' denotes unavailable results.
}
\label{tab:rstpreid_comparison}

\begin{tabular}{llcccccc}
\toprule

\multirow{2}{*}{\textbf{Method}} &
\multirow{2}{*}{\textbf{Venue}} &
\multicolumn{3}{c}{\textbf{Supervised}} &
\multicolumn{3}{c}{\textbf{Zero-shot Transfer}}\\
\cmidrule(lr){3-5}
\cmidrule(lr){6-8}
&&

\textbf{R@1} & \textbf{R@5} & \textbf{R@10} &
\textbf{R@1} & \textbf{R@5} & \textbf{R@10} \\
\midrule

DSSL~\cite{zhu2021dssl}
& MM'21
& 32.43 & 55.08 & 63.19
& -- & -- & -- \\

IVT~\cite{shu2022see}
& ECCVW'22
& 46.70 & 70.00 & 78.80
& 43.70 & 65.10 & 75.55 \\

CFine~\cite{yan2023clip}
& TIP'23
& 50.55 & 72.50 & 81.60
& 47.40 & 70.60 & 79.35 \\

IRRA~\cite{jiang2023cross}
& CVPR'23
& 60.20 & 81.30 & 88.20
& 45.10 & 69.20 & 78.75 \\

RDE~\cite{qin2024noisy}
& CVPR'24
& 65.35 & 83.95 & 89.90
& -- & -- & -- \\

APTM~\cite{yang2023towards}
& MM'23
& \textbf{67.50} & \textbf{85.70} & \textbf{91.45}
& 52.50 & \textbf{75.15} & \textbf{81.70} \\

RaSa~\cite{yang2023rasa}
& IJCAI'23
& -- & -- & --
& 55.00 & 73.65 & 81.55 \\
\midrule
CMP~\cite{yang2025beyondwalking}
& ICCV'25
& -- & -- & --
& 29.15 & 50.40 & 60.75 \\

\rowcolor{blue!5}
\textbf{ActPair (Ours)}
& ECCVW'26
& -- & -- & --
& \textbf{55.25} & 69.05 & 73.75 \\

\bottomrule
\end{tabular}

\vspace{-4mm}
\end{table}

\textbf{Main Results on RSTPReid} 
To assess the robustness and transferability of the learned representations in general, non-anomaly-specific environments, Table~\ref{tab:rstpreid_comparison} reports the performance on the RSTPReid dataset~\cite{zhu2021dssl}. The evaluation is divided into supervised fine-tuning and cross-dataset evaluation scenarios. Under the cross-dataset setting, where models are evaluated without utilizing any RSTPReid~\cite{zhu2021dssl} samples for training or domain adaptation, ActPair substantially improves over the same-source CMP baseline. Specifically, when transferring directly from the PAB dataset to RSTPReid using the full pipeline with pairwise reranking, our approach yields a Recall@1 of 55.25\%, a Recall@5 of 69.05\%, and a Recall@10 of 73.75\%. This constitutes a substantial improvement over the baseline CMP model~\cite{yang2025beyondwalking}, which attains only 29.15\% Recall@1 under identical zero-shot transfer conditions; since reranking only reorders the top-10, the lower Recall@5/@10 relative to CFine, APTM, and IRRA reflects the retrieval stage's coverage rather than a reranking limitation.

While RSTPReid predominantly relies on appearance rather than behavioral anomalies, ActPair substantially improves over the same-source CMP baseline without using RSTPReid annotations or domain adaptation. This suggests that the action-aligned representations learned on PAB retain useful cross-domain information, particularly for top-1 retrieval.

\vspace{-2mm}
\subsection{Qualitative Analysis}
\vspace{-2mm}
\begin{figure}[!t]
    \centering
    \includegraphics[width=\linewidth]{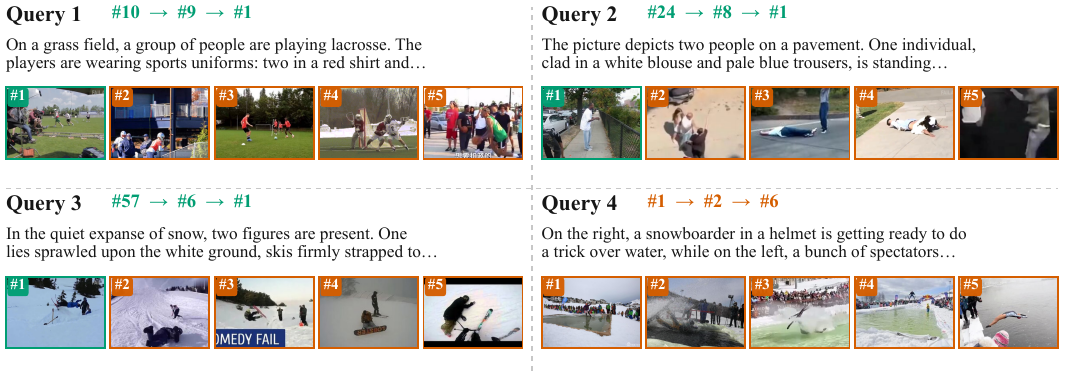}
    \caption{Qualitative top-5 retrieval results across the three stages. The numbers indicate the ground-truth ranks after action-aligned retrieval, dual-view query fusion, and pairwise reranking. Queries~1--3 are successful cases, while Query~4 illustrates a failure on complex spatial and action-state constraints.}
    \label{fig:qualitative}
    \vspace{-4mm}
\end{figure}

Figure~\ref{fig:qualitative} shows representative retrieval results across the three stages. For queries 1–3, dual-view fusion improves candidate discovery while pairwise reranking promotes the ground-truth image to rank 1; Query 3 illustrates this clearly, moving from rank 57 to 6 to 1 across the three stages. Query~4 presents a failure case: although the correct image is initially ranked first, it is demoted by the later stages. This suggests that the model may still struggle with long compositional descriptions involving action phases, multiple people, and precise spatial relations.

\vspace{-2mm}
\section{Ablation Study}

\vspace{-2mm}

\subsection{Contribution of Each Stage}
Table~\ref{tab:two_set_stage_ablation} presents a progressive evaluation of the framework's components on the Challenge and PAB Public Test sets. The baseline configuration, employing the action-supervised SigLIP2 dual-encoder, establishes a strong foundation with a Recall@1 of 72.14\% on the Challenge Set and 84.98\% on the Public Test set. This configuration outperforms the pose-aware CMP baseline, indicating higher initial retrieval accuracy under the evaluated setting. Building upon the action-aligned retriever, dual-view query fusion preserves fine-grained query details while leveraging the context-grounded rewrite. This parallel retrieval strategy captures complementary semantic evidence, improving Recall@1 to 75.73\% and 85.95\% on the respective test sets. Finally, incorporating the training-free pairwise multimodal reranking further improves top-rank retrieval among the top-10 candidates. This complete framework delivers the most substantial performance gain, elevating the Challenge Set Recall@1 to 84.88\% and achieving a final Recall@1 of 88.62\% alongside a 93.97\% mAP on the Public Test set. These progressive improvements demonstrate the complementary contribution of each stage in isolating fine-grained behavioral anomalies.

\begin{table*}[t]
\centering
\caption{Progressive evaluation on the AI City 2026 challenge test
and the original PAB test.
}
\label{tab:two_set_stage_ablation}

\renewcommand{\arraystretch}{1.12}

\begin{tabular}{lcccc|cccc}
\toprule
\multirow{2}{*}{\textbf{Method}}
& \multicolumn{4}{c|}{\textbf{Challenge Set}}
& \multicolumn{4}{c}{\textbf{Public Test}} \\
\cmidrule(lr){2-5}
\cmidrule(lr){6-9}
& \textbf{R@1} & \textbf{R@5} & \textbf{R@10} & \textbf{mAP}
& \textbf{R@1} & \textbf{R@5} & \textbf{R@10} & \textbf{mAP} \\
\midrule

CMP~\cite{yang2025beyondwalking}
& 69.41 & 89.43 & 91.91 & --
& 84.93 & 99.09 & 99.75 & 91.66 \\

\midrule

Action-Aligned Retriever
& 72.14 & 92.77 & 95.50 & 81.68
& 84.98 & 99.29 & 99.60 & 91.73 \\

+ Dual-View Query
& 75.73 & 94.99 & \textbf{97.27} & 84.48
& 85.95 & 99.34 & \textbf{99.85} & 92.30 \\

+ Pairwise Reranking
& \textbf{84.88} & \textbf{96.36} & \textbf{97.27} & \textbf{90.14}
& \textbf{88.62} & \textbf{99.75} & \textbf{99.85} & \textbf{93.97} \\

\bottomrule
\end{tabular}
\end{table*}

\begin{table}[h]
\centering
\caption{
Controlled comparison of pose- and action-based supervision on the challenge set. The SigLIP2~\cite{tschannen2025siglip} comparison keeps the backbone and training configuration fixed, isolating the effect of the supervision signal.
}
\label{tab:action_pose_ablation}
\begin{tabular}{llccc}
\toprule
\textbf{Backbone} &
\textbf{Supervision} &
\textbf{R@1} &
\textbf{R@5} &
\textbf{R@10} \\
\midrule

\rowcolor{black!4}
\multicolumn{5}{l}{\textit{Effect of pose information in CMP~\cite{yang2025beyondwalking}}} \\
CMP & RGB and text                    & 68.60 & 87.51 & 89.64 \\
CMP & RGB, text, and pose             & 69.41 & 89.43 & 91.91 \\
\addlinespace[2pt]
\midrule
\rowcolor{blue!5}
\multicolumn{5}{l}{\textit{Controlled comparison under the same SigLIP2 backbone~\cite{tschannen2025siglip}}} \\
SigLIP2 & Pose-based supervision       & 67.95 & 91.20 & 94.74 \\
SigLIP2 & \textbf{Action-based supervision} &
\textbf{72.14} & \textbf{92.77} & \textbf{95.50} \\

\multicolumn{2}{l}{\quad $\Delta$ Action $-$ Pose \textit{Paired bootstrap 95\% CI}} &
\textbf{+4.19} &
\textbf{+1.57} &
+0.76 \\
&
&
{\scriptsize [+$2.38$, +$6.07$]} &
{\scriptsize [+$0.51$, +$2.63$]} &
{\scriptsize [$-0.10$, +$1.62$]} \\
\bottomrule
\end{tabular}
\end{table}

\vspace{-2mm}
\subsection{Action versus Pose Supervision}

\vspace{-2mm}
Table~\ref{tab:action_pose_ablation} isolates the effectiveness of explicit action supervision relative to skeletal pose signals. Incorporating pose information into the CMP baseline yields only a modest improvement, increasing Recall@1 from 68.60\% to 69.41\%. However, under a controlled comparison using the same SigLIP2 backbone~\cite{tschannen2025siglip}, pose-based supervision remains less effective than our explicit action supervision. Employing human pose extracted via RTMPose~\cite{jiang2023rtmpose} as the primary training signal proved suboptimal, achieving only 67.95\% Recall@1. This empirically highlights the pose-semantic gap: isolated skeletal structures fail to uniquely define actions without environmental context.  Conversely, applying explicit action-tag supervision to the SigLIP2 backbone significantly enhances retrieval accuracy, yielding 72.14\% Recall@1, 92.77\% Recall@5, and 95.50\% Recall@10. Paired bootstrap analysis confirms a statistically significant improvement of +4.19\% in Recall@1 (95\% CI: [+2.38, +6.07]) for action-based over pose-based supervision. These results demonstrate that optimizing for context-conditioned action semantics is substantially more discriminative for anomaly search than relying on geometric constraints.  

\vspace{-2mm}
\subsection{Pointwise versus Pairwise Reranking}
\label{ab:pair_wise}
\vspace{-2mm}
Table~\ref{tab:pointwise_pairwise} evaluates the proposed pairwise reranking strategy against a pointwise scoring baseline using the identical Qwen3.5-9B model and candidate pool. Assigning an absolute, independently calibrated scalar score (pointwise) requires the model to maintain a consistent scale across unrelated queries. In contrast, pairwise comparison simplifies the reasoning objective by tasking the model solely with determining which of two simultaneously observed candidates better satisfies the specific query criteria.  

Empirically, while pointwise VLM reranking improves the initial dense retrieval baseline (increasing Recall@1 from 75.73\% to 80.13\%), the pairwise strategy is substantially more discriminative. It achieves 84.88\% Recall@1 and 90.14\% mAP, outperforming the pointwise approach by absolute margins of +4.75\% and +2.44\%, respectively (95\% CI for Recall@1: [+3.08, +6.42]). Specifically, this ranking effectiveness substantially reduces the number of comparisons relative to exhaustive all-pairs reranking. Under our top-10 configuration, pairwise reranking uses an average of 9.09 model calls per query, compared with 10.00 calls for the pointwise implementation.

\begin{table}[t]
\centering
\caption{
Comparison of pointwise and pairwise VLM reranking on the challenge set. Both variants use the same Qwen3.5-9B model~\cite{qwen35blog} and candidate pool. Requests denote the average number of VLM calls per query.
}
\label{tab:pointwise_pairwise}
\begin{tabular}{lccccc}
\toprule
\textbf{Reranking strategy} &
\textbf{R@1 $\uparrow$} &
\textbf{R@5 $\uparrow$} &
\textbf{R@10 $\uparrow$} &
\textbf{mAP $\uparrow$} &
\textbf{Req./Q $\downarrow$} \\
\midrule
No VLM reranking
& 75.73 & 94.99 & 97.27 & 84.48 & 0.00 \\

Qwen3.5~\cite{qwen35blog} pointwise
& 80.13 & \textbf{96.41} & 97.27 & 87.70 & 10.00 \\

\textbf{Qwen3.5~\cite{qwen35blog} pairwise}
& \textbf{84.88} & 96.36 & 97.27 & \textbf{90.14} & \textbf{9.09} \\

\midrule
Pairwise $-$ Pointwise
& \textbf{+4.75}
& $-0.05$
& 0.00
& \textbf{+2.44}
& $-$\textbf{0.91} \\

\scriptsize \textit{Paired bootstrap 95\% CI}
& \scriptsize [+$3.08$, +$6.42$]
& \scriptsize [$-0.61$, +$0.51$]
& \scriptsize [0, 0]
& \scriptsize [+$1.45$, +$3.43$]
& -- \\
\bottomrule
\end{tabular}
    \vspace{-4mm}
\end{table}
\vspace{-2mm}
\section{Conclusion}

\vspace{-2mm}
In this paper, we propose a three-stage framework for text-based person anomaly search that bridges the pose-semantic gap via action-aligned representation learning, context-grounded query fusion, and training-free pairwise multimodal reranking. ActPair achieves the best performance among the compared published baselines on the original PAB test and substantially improves over the same-source CMP baseline in cross-dataset transfer to RSTPReid. Beyond pedestrian anomaly search, ActPair may support other context-grounded retrieval tasks, such as distinguishing similar actions in sports. We view our framework as a step toward such broader context-grounded retrieval tasks.

\textbf{Limitations and Future Work}: Constructing canonical vocabularies ($\mathcal{T}_{act}$ and $\mathcal{T}_{scene}$) currently requires dataset-specific annotations, limiting scalability in unannotated domains. Future work will eliminate this dependency by utilizing an unsupervised MLLM pipeline to dynamically parse raw captions into action and scene clusters.

\textbf{Acknowledgement.} This work was supported by Saigon AI Hub, jointly established by VNG Group JSC and Vietnam National University Ho Chi Minh City (VNU-HCM), which provided computing infrastructure, research resources, and a collaborative research environment. Trong-Thuan Nguyen was funded by the PhD Scholarship Programme of Vingroup Innovation Foundation (VINIF), VinUniversity, code VINIF.2025.TS63.

\bibliographystyle{splncs04}
\bibliography{main}
\end{document}